\documentclass[runningheads]{llncs}

\usepackage[T1]{fontenc}
\usepackage{float}
\usepackage{graphicx}
\usepackage{amsmath}
\usepackage{soul}
\usepackage{xcolor}
\usepackage{hyperref}

\definecolor{myblue}{rgb}{0,0,0.7}

\definecolor{myred}{rgb}{0.7,0,0}

\definecolor{mygreen}{rgb}{0,0.4,0}

\definecolor{myorange}{rgb}{1.0,0.4,0}

\usepackage{lineno}

\begin{document}

\title{Upgrading Pepper Robot's Social Interaction with Advanced Hardware and Perception Enhancements}
\titlerunning{Upgrading Pepper Robot's Social Interaction}

\author{Paolo Magri \and
Javad Amirian \and
Mohamed Chetouani}
\authorrunning{Paolo Magri, et al.}

\institute{Sorbonne Université, Institut des Systèmes Intelligents et de Robotique (ISIR)\\ Paris, France \\
\email{[magri, amirian, chetouani]@isir.upmc.fr}
}

\maketitle

\begin{abstract}

In this paper, we propose hardware and software enhancements for the Pepper robot to improve its human-robot interaction capabilities.
This includes the integration of an NVIDIA Jetson GPU to enhance computational capabilities and execute real-time algorithms, and a RealSense D435i camera to capture depth images,
as well as the computer vision algorithms to detect and localize the humans around the robot and estimate their body orientation and gaze direction.
The new stack is implemented on ROS and is running on the extended Pepper hardware, and the communication with the robot's firmware is done through the NAOqi ROS driver API.
We have also collected a MoCap dataset of human activities in a controlled environment, together with the corresponding RGB-D data, to validate the proposed perception algorithms.
The repository with the dataset, the codes and the 3D hardware CAD is available at \url{github.com/polmagri/Enhancing-Pepper-Robot-s-HRI}.


\end{abstract}

\section{Introduction}



%

The integration of social robots into daily environments has advanced significantly, driving the need for improved human-robot interaction (HRI).
This paper builds on the existing body of work \cite{amirian2024legibot} \cite{caniot_et_al} by focusing on the Pepper robot, a widely used social robot, and explores both hardware and software enhancements to improve its interaction capabilities.
Detailed hardware upgrades are discussed in section 2.
These enhancements are designed to support advanced object detection and pose estimation algorithms, along with the calculation of human gaze and body orientation, thus enabling more effective HRI in real-world scenarios.
Comparing our work with the HRI ROS package proposed by Mohamed et al. \cite{mohamed_et_al},
the software in our stack are designed to run on GPUs and the integrated Jetson computer on Pepper has less impact than installing a laptop on the robot.
Our core contributions include:
\begin{enumerate}
    \item Detailed instructions for implementing hardware upgrades on the Pepper robot, including the installation of an NVIDIA Orin Jetson Nano GPU and a RealSense RGBD camera.
    \item Integration of ROS-based software for human detection and field-of-view estimation with the robot's firmware, resulting in a flexible system.
    \item The development of a MoCap dataset for testing and improving the robot's perception algorithms.
\end{enumerate}



\section{Hardware Upgrades to the Pepper Robot}



Pepper as one of the most popular social robots, has been widely used in various applications, including healthcare, education, and entertainment.
However, the robot's hardware limitations have hindered its ability to perform advanced tasks, and it can no longer meet the demands of modern deep learning algorithms.
Its processing performance is insufficient to execute real-time perception and with its perception hardware, it would not be able to capture social cues effectively.


To overcome these limitations, we implemented several hardware upgrades to the robot.
First, we have added an NVIDIA Jetson Orin Nano GPU to enhance the robot's computational capabilities.
The AI performance of the new CPU is significantly higher than the original, as shown in Table \ref{tab:comparison}.
The choice of the Jetson Orin Nano was made due to a trade-off between several factors: the Jetson's high computational power, its small form factor making it suitable to mount on the robot, its low power consumption, and its affordable price.
%
As a secondary enhancement, we have integrated the Intel RealSense D435i camera into the Pepper robot's hardware configuration.
Given budget constraints and the requirements of our applications, the RealSense presents an optimal balance between cost and performance.
This camera is pivotal for enabling 3D depth estimation providing detailed depth information and it is particularly beneficial for the robot's interaction with its surroundings.
Finally, an external battery is added, to power the new hardware for at least 70 minutes \cite{amirian2024legibot}, as there is no easy way to power the new hardware from the robot's internal battery.

\vspace{-0.4cm}
\begin{table}
\centering
\caption{Summary of Hardware Upgrades}\label{tab:comparison}
\begin{tabular}{|p{2cm}|p{3.5cm}|p{6.3cm}|}
\hline
\textbf{\centering Item} & \textbf{\centering Original } & \textbf{\centering Extension} \\
\hline
\textbf{CPU} & \begin{tabular}[c]{@{}l@{}}Intel Atom E3845, \\ Quad-core 1.91 GHz\end{tabular} & \begin{tabular}[c]{@{}l@{}}Arm Cortex-A78AE,\\ 6-core 1.5 GHz\end{tabular} \\
\hline
\textbf{GPU} & None & \begin{tabular}[c]{@{}l@{}}NVIDIA Jetson Orin Nano,\\ 1024 CUDA cores, 32 tensor cores, 625MHz\end{tabular} \\
\hline
\textbf{AI Performance} & \begin{tabular}[c]{@{}l@{}}51 GFLOPS (FP32)\end{tabular} & \begin{tabular}[c]{@{}l@{}} 40 TOPS (INT8)\end{tabular} \\
\hline
\textbf{Camera} & \begin{tabular}[c]{@{}l@{}}RGB 30 fps - 640x480\end{tabular} & \begin{tabular}[c]{@{}l@{}} RGBD 30 fps - 1280x720\end{tabular} \\
\hline
\textbf{Battery} & \begin{tabular}[c]{@{}l@{}}29.2v, 30,000 mAh\end{tabular} & \begin{tabular}[c]{@{}l@{}}11.1v, 2200 mAh\end{tabular} \\
\hline
\end{tabular}
\end{table}
\vspace{-0.6cm}
%
%
\subsubsection{CAD Design for Hardware Integration:}
To accommodate the external GPU, camera and battery, we have designed customized 3D printed parts.
the structural modifications have been optimized to withstand the operational demands placed on the robot, ensuring that it can perform its duties without any hardware-related disruptions.
\vspace{-10pt}
\begin{figure}[ht]
    \centering
    \begin{minipage}{1\textwidth}
        \centering
        \includegraphics[width=\textwidth]{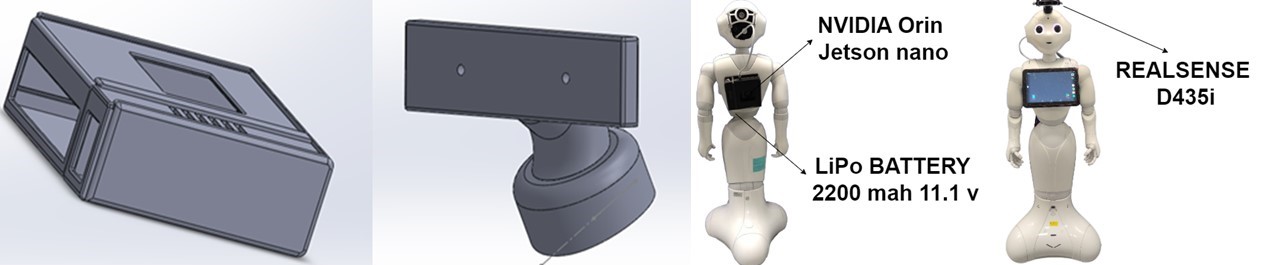}
    \end{minipage}
    \caption{1) Customized box to house the GPU and the battery separated by a partition. It is equipped with air vents to ensure adequate cooling, essential for maintaining optimal performance and longevity of the GPU. 2) The camera mount is developed sturdy and compact and also thick and short to minimize the vibrations. Due to the degrees of freedom of the Pepper's head, there is no need for more complex mechanisms, and given the fixed geometry of the mount, the camera can be easily calibrated. 3) Back Pepper Robot with New Hardware Installed 4) Front Pepper Robot with New Hardware Installed  }
\end{figure}
\vspace{-10 pt}


\section{Human Detection and FOV Estimation}


Even though Pepper is equipped with human tracking capabilities, but it is unable to localize the 3D position of humans and estimate their field of view (FOV) and body orientation which are crucial for social robots to understand the human's intentions.
Here we propose the software components to enhance Pepper's perception capabilities, including human detection, pose estimation, and FOV estimation.
This stack is implemented on ROS noetic, as the ROS2 driver for Pepper is still unstable.

\begin{figure}[ht]
    \centering
    \begin{minipage}{1\textwidth}
        \centering
        \includegraphics[width=\textwidth]{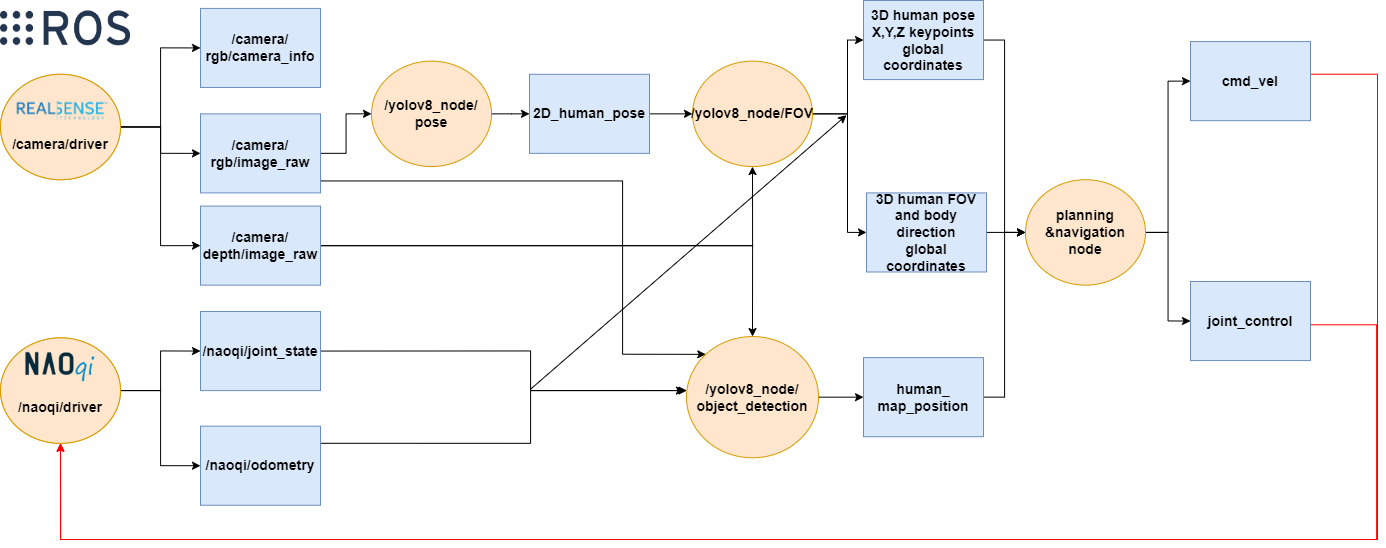}
    \end{minipage}
    \caption{ROS stack for human detection and FOV estimation}
\end{figure}
\vspace{-10 pt}

\subsubsection{YOLOv8 for Human Detection and Pose Estimation:}
The YOLOv8-pose variant excels in real-time human detection and body keypoint estimation. Leveraging the efficient architecture of YOLOv8, the quantized model (using TensorRT) achieves a performance of 30 fps, matching the frame rate of the RealSense camera, making it well-suited for social interaction scenarios.

\subsubsection{3D Keypoint Localization and Depth Integration:}
The YOLOv8-pose model generates 2D coordinates for human keypoints, which are then combined with depth information from the RealSense camera to determine the corresponding 3D coordinates of keypoints such as eyes, shoulders, and knees.
The fusion of 2D and 3D data presents challenges, necessitating the development of algorithms to enhance the accuracy of 3D localization.

\begin{itemize}
    \item Background and Foreground Distinction: For each human keypoint, a circle with a radius of 5 is calculated, and the minimum depth value within this circle is selected to minimize noise and measurement discrepancies.
    \item Correction Algorithms: The Kalman Filter is employed to improve the consistency and accuracy of field-of-view (FOV) estimation, particularly in critical situations such as when a person is in profile relative to the camera.
\end{itemize}

\subsubsection{Estimating Body Orientation and Gaze Direction:}
The orientation of the torso for each frame is defined by calculating the average vector between the keypoints of the shoulders and hips on a horizontal plane. Similarly, the direction of gaze is determined based on the keypoints of the eyes and neck. \\
\begin{figure}[ht]
    \centering
    \begin{minipage}{1\textwidth}
        \centering
        \includegraphics[width=\textwidth]{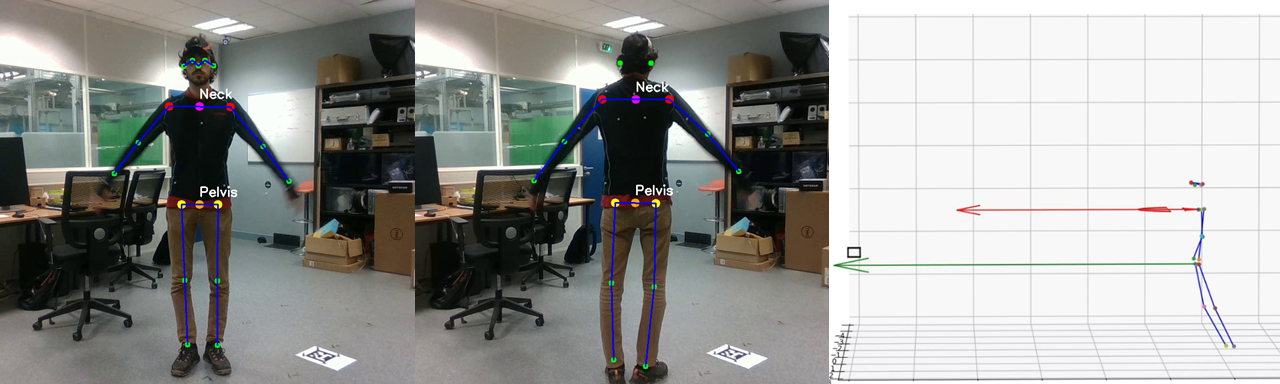}
    \end{minipage}
    \caption{Left) Subject facing the camera. Center) Subject with back to the camera. Right) Side view of the 3D skeleton. \\
    \textbf{Legend:} Red: \(\mathbf{P}_\text{Shoulders}\), Yellow: \(\mathbf{P}_\text{Hips}\), Orange: \(\mathbf{P}_\text{Pelvis}\), Pink: \(\mathbf{P}_\text{Neck}\). \\
    Red arrow: \(\mathbf{q}_\text{gaze}\), Green arrow: \(\mathbf{q}_\text{torso}\).}
    \label{fig:algor}
\end{figure}
\noindent
\\
1) The normal vector \( \mathbf{N} \) to the plane formed by the cross product of the shoulder-hip and shoulder-shoulder vectors is calculated as follows:
\begin{equation}
\mathbf{N}_\text{torso} = (\mathbf{P}_\text{shoulder.L} - \mathbf{P}_\text{pelvis}) \times (\mathbf{P}_\text{shoulder.R} - \mathbf{P}_\text{pelvis})
\end{equation}
\begin{equation}
\mathbf{N}_\text{gaze} = (\mathbf{P}_\text{eye.L} - \mathbf{P}_\text{neck}) \times (\mathbf{P}_\text{eye.R} - \mathbf{P}_\text{neck})
\end{equation}

\noindent
2) The projection of the normal vector onto the XY-plane and the normalized direction vectors for the torso and the gaze are:
\begin{equation}
\mathbf{d}_\text{torso} = \frac{\mathbf{N}_{\text{torso}, xy}}{\|\mathbf{N}_{\text{torso}, xy}\|}, \quad
\mathbf{d}_\text{gaze} = \frac{\mathbf{N}_{\text{gaze}, xy}}{\|\mathbf{N}_{\text{gaze}, xy}\|}
\end{equation}

\noindent
3) The quaternion representing the rotation from the forward direction \([1, 0, 0]\) to the body direction vector and the view direction vector is:
\footnotesize{
\begin{equation}
\mathbf{q}_\text{torso} = \left[ \cos\left(\frac{\theta_\text{torso}}{2}\right), \mathbf{d}_{\text{torso}, x} \sin\left(\frac{\theta_\text{torso}}{2}\right), \mathbf{d}_{\text{torso}, y} \sin\left(\frac{\theta_\text{torso}}{2}\right), \mathbf{d}_{\text{torso}, z} \sin\left(\frac{\theta_\text{torso}}{2}\right) \right]
\end{equation}
\begin{equation}
\mathbf{q}_\text{gaze} = \left[ \cos\left(\frac{\theta_\text{gaze}}{2}\right), \mathbf{d}_{\text{gaze}, x} \sin\left(\frac{\theta_\text{gaze}}{2}\right), \mathbf{d}_{\text{gaze}, y} \sin\left(\frac{\theta_\text{gaze}}{2}\right), \mathbf{d}_{\text{gaze}, z} \sin\left(\frac{\theta_\text{gaze}}{2}\right) \right]
\end{equation}
\normalsize
}

\noindent

We adopted a 120-degree horizontal human FOV in accordance with \cite{taylor_et_al}, to estimate the human's FOV.

\section{MoCap Dataset}


The objective of this dataset is to capture keypoints of a person using a motion capture (MoCap) system in a controlled environment with standard lighting, and to use as a baseline for testing the proposed FOV estimation stack.
The dataset contains various types of human movements, capturing different distances, orientations, heights, and poses.
Each trial lasts approximately 2 minutes, featuring a single person in the room.
The algorithm is assumed to function in multi-person scenarios, without considering the loss of information due to occlusions by objects or other people.
\noindent
The data acquisition setup includes:
\begin{itemize}
\item MoCap System: 8 cameras arranged to cover the entire room, ensuring comprehensive capture of human movements.
\item Realsense Camera Positioned statically at a height of 1.25 meters, capturing scenes at 1280x720 pixels and 30 fps, matching Pepper's upright view.
\item Synchronization: Spatial calibration using Apriltags and temporal synchronization using ROS time.
\item Data Storage Format: Captured data is stored in ROS bags, containing RGB-D images, MoCap skeleton outputs, and the corresponding timestamps.
\end{itemize}

\noindent
The participants are instructed to perform the following activities:
\begin{itemize}
\item Standard walk back and forth
\item Walk back and forth with arms crossed
\item Walk back and forth with sudden movements (e.g., dodging an obstacle)
\item Walk back and forth in a zigzag pattern with pronounced head movements
\end{itemize}

This dataset is used to correct the errors of the algorithm by applying filters to avoid mistakes when the person is in profile or turned away from the camera (difficulty in detecting keypoints). The MoCap data is used as ground truth, while the Realsense data simulates its application on Pepper.
\begin{figure}[ht]
    \centering
    \begin{minipage}{1\textwidth}
        \centering
        \includegraphics[width=\textwidth]{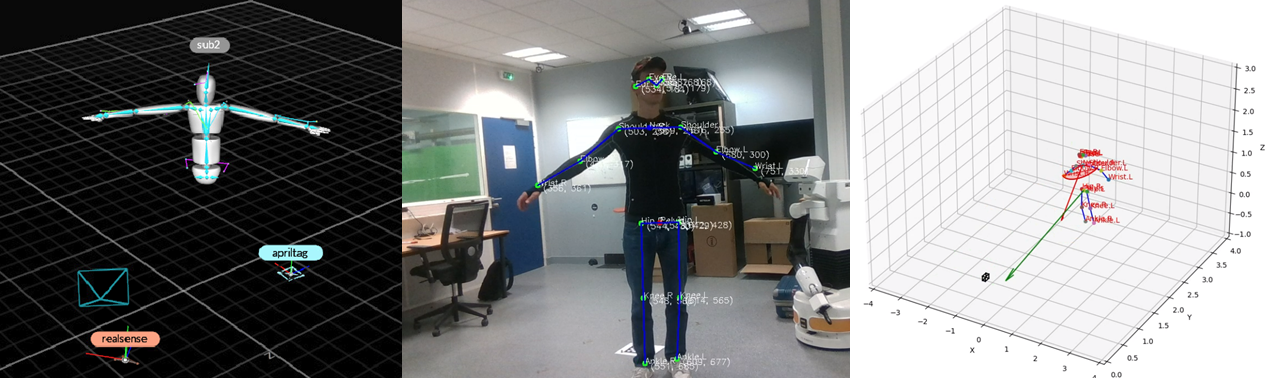}
    \end{minipage}
    \caption{Left) Skeleton generated from MoCap. Middle) RGB image from RealSense with YOLOv8 pose. Right) skeleton with FOV (red) and body direction (green) generated by algorithm}
\end{figure}
\vspace{-2.4pt }

\section{Conclusion} 
In this paper, we have proposed hardware and software enhancements for the Pepper robot to improve its human-robot interaction capabilities.
The hardware upgrades include the integration of an NVIDIA Jetson GPU and a RealSense camera, while the software enhancements involve human detection, pose estimation, and FOV estimation.
And it can be easily installed on the Pepper robot, with a total cost of approximately 800 euros.
The software stack is implemented on ROS, and makes it possible to integrate with Pepper's firmware and also other ROS software.
 \section{Acknowledgments}
This work has received funding from the European Union’s Horizon 2020 research
and innovation programmes under grant agreement No 952026 and Horizon Europe
Framework Programme under grant agreement No 101070596.


%
%
%

\end{document}